\documentclass{article}

\PassOptionsToPackage{numbers, compress}{natbib}

    \usepackage[final]{neurips_2020}

\usepackage[utf8]{inputenc} 
\usepackage[T1]{fontenc}    
\usepackage{hyperref}       
\usepackage{url}            
\usepackage{booktabs}       
\usepackage{amsfonts}       
\usepackage{nicefrac}       
\usepackage{microtype}      
\usepackage{amssymb}
\usepackage{multirow}
\usepackage{comment}
\usepackage{graphicx}
\usepackage{amsthm}
\usepackage{amsmath}
\usepackage{wrapfig}
\usepackage[table]{xcolor}
\usepackage{color}
\usepackage{breqn}
\usepackage{bbm}
\newtheoremstyle{mystyle}
    {}
    {}
    {}
    {}
    {\bf}
    {.}
    { }
    {\underline{\thmname{#1}\thmnumber{#2}\thmnote{（#3）}}}%
\theoremstyle{mystyle}

\newtheorem*{theorem*}{Theorem}

\newcommand{\bhline}[1]{\noalign{\hrule height #1}}
\newcommand{\specialcell}[2][c]{%
  \begin{tabular}[#1]{@{}c@{}}#2\end{tabular}}
\title{Neural Star Domain as Primitive Representation}

\author{%
  \textbf{Yuki Kawana$^1$, Yusuke Mukuta$^{1,2}$, Tatsuya Harada$^{1,2}$}\\
  $^{1}$The University of Tokyo, $^{2}$RIKEN AIP\\
  \texttt{\{kawana, mukuta, harada\}@mi.t.u-tokyo.ac.jp}
}

\begin{document}

\maketitle

\begin{abstract}
Reconstructing 3D objects from 2D images is a fundamental task in computer vision. Accurate structured reconstruction by parsimonious and semantic primitive representation further broadens its application. When reconstructing a target shape with multiple primitives, it is preferable that one can instantly access the union of basic properties of the shape such as collective volume and surface, treating the primitives as if they are one single shape. This becomes possible by primitive representation with unified implicit and explicit representations. However, primitive representations in current approaches do not satisfy all of the above requirements at the same time. To solve this problem, we propose a novel primitive representation named neural star domain (NSD) that learns primitive shapes in the star domain. We show that NSD is a universal approximator of the star domain and is not only parsimonious and semantic but also an implicit and explicit shape representation. We demonstrate that our approach outperforms existing methods in image reconstruction tasks, semantic capabilities, and speed and quality of sampling high-resolution meshes. 
\end{abstract}

\section{Introduction}

Understanding 3D objects by decomposing them into simpler shapes, called primitives, has been widely studied in computer vision \cite{roberts1963machine, binford1971visual, biederman1987recognition}. Decomposing 3D objects into parsimonious and semantic primitive representations is important to understand their structures. Constructed solid geometry \cite{laidlaw1986constructive} is a field of research using primitives, where complex shapes are reconstructed by combinations such as union of primitives. 

Recently, learning based approaches have been adopted in the above fields of  studies \cite{chen2019bsp, deng2019cvxnets, deprelle2019learning, Paschalidou2019CVPR, Paschalidou2020CVPR, niu2018im2struct, tulsiani2017learning}. It has been shown that learning based approaches enable semantically consistent part arrangement in various shapes. Moreover, the use of implicit representations allows the set of primitives to be represented as a single collective shape by considering a union \cite{chen2019bsp, deng2019cvxnets, genova2019learning}. This property contributes to improving the reconstruction accuracy during training.

However, the expressiveness of primitives, especially with closed shapes, has been limited to simple shapes (cuboids, superquadrics, and convexes) in existing studies. Although the primitives can learn semantic part arrangements, semantic shapes of the parts cannot be learned in existing methods. In addition, although the union of primitive volumes could be represented by implicit representations in previous studies, lack of instant access to the union of primitive surfaces during training result in complex training schemes \cite{chen2019bsp, deng2019cvxnets, genova2019learning}.

It is challenging to define a primitive that addresses all of these problems. State-of-the-art expressive primitives with explicit surfaces do not have implicit representations \cite{groueix2018papier, deprelle2019learning}, and thus, are unable to efficiently consider unions of primitives to represent collective shapes. Leading primitive representations by convexes \cite{chen2019bsp, deng2019cvxnets} with implicit representations have a trade-off on the number of half-space hyperplanes H consisting a convex. Using more hyperplanes yield more expressive convexes at the expense of a quadratically growing computation cost in extracting differentiable surface points. A naive implementation costs $O(H^2)$ to filter only the surface points of a convex from the hyperplanes. 

\begin{figure}

  \centering
      \includegraphics[width=13cm]{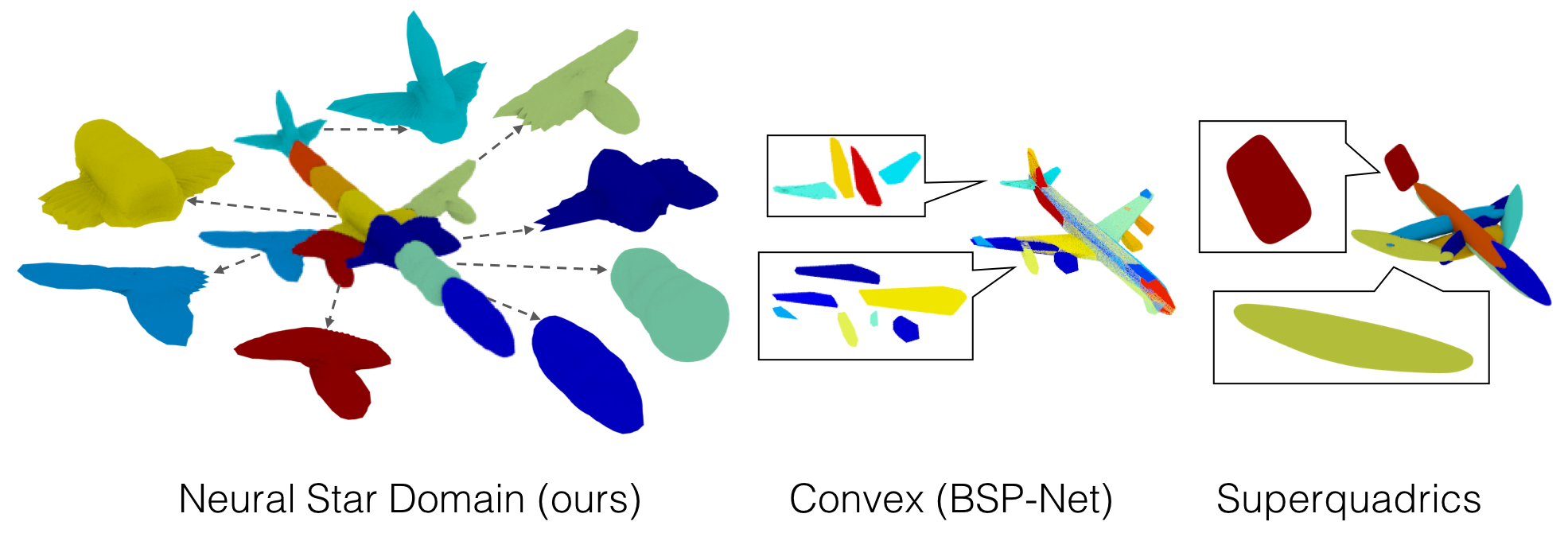}
              \caption{Overview of our approach. The proposed primitives have more meaningful and wider variety of shapes compared to previous works.}
        \vspace{-1\baselineskip}

\label{fig:overview}
\end{figure}

To address these issues, we propose a novel primitive representation named neural star domain (NSD) that learns shapes in the star domain by neural networks. Star domain is a group of arbitrary shapes that can be represented by a continuous function defined on the surface of a sphere. As it can express concavity, we can regard it as a generalized shape representation of convexes. We visualize the learned primitives in Figure \ref{fig:overview}. Moreover, we can directly approximate star domain shapes with neural networks due to their continuities. We demonstrate that the complexity of shapes the neural star domain can represent is equivalent to the approximation ability of the neural network. In addition, as it is defined on the surface of sphere, we can represent the primitive in both implicit and explicit forms by transforming it between spherical and Cartesian coordinates. We compare our approach to previous studies in Table \ref{tb:comp_agains_prevworks}.

Our contributions can be summarized as follows: (1) We propose a novel primitive representation called NSD having high expressive power. We show that this novel primitive representation is more parsimonious and is able to learn semantic part shapes. (2) We show that our proposed primitive provides unified implicit and explicit representations that can be used during training and inference, leading to improved mesh reconstruction accuracy and speed.

\begin{table}
\centering

\begin{tabular}{c|ccccc}
\bhline{1.5 pt}
\tabcolsep = 1pt
& Implicit 
& Explicit
& Semantic
& Parsimonious 
& Accurate
\\ \hline 
DMC \cite{liao2018deep} & \checkmark & \checkmark & -- & --  & \checkmark  \\
SQ \cite{Paschalidou2019CVPR} & & \checkmark &   & \checkmark & \\
AtlasNetV2 \cite{deprelle2019learning} & & \checkmark &\checkmark  & \checkmark  & \checkmark   \\
BSP-Net \cite{chen2019bsp} & \checkmark  &   &   &   &  \checkmark \\ 
Ours & \checkmark & \checkmark &\checkmark & \checkmark  & \checkmark \\ 
\bhline{1.5 pt}
\end{tabular}
\vspace{+1\baselineskip}
\caption{Overview of shape representation in previous works. SQ denotes superquadrics \cite{Paschalidou2019CVPR}. We regard a primitive as having an explicit representation if it has access to the explicit surface in \emph {both} the inference and the training. Moreover, we say that a primitive representation is semantic if it can reconstruct semantic \emph{shapes} in addition to part correspondence.}
        \vspace{-1\baselineskip}
\label{tb:comp_agains_prevworks}
\end{table}


\section{Related works}
Methods to decompose shapes to primitives have been studied extensively in computer vision \cite{roberts1963machine}. Some of the classical primitives used in computer vision are generalized cylinders \cite{binford1971visual} and geons \cite{biederman1987recognition}. For deep generative models, cuboids \cite{tulsiani2017learning, niu2018im2struct} and superquadrics \cite{Paschalidou2019CVPR, Paschalidou2020CVPR} are used to realize consistent parsing across shapes. However, these methods have poor reconstruction accuracies due to limitations in the parameter spaces of the primitives. Thus, their application is limited to shape abstraction. Using parametrized convexes to improve the reconstruction accuracy has been recently proposed in \cite{chen2019bsp, deng2019cvxnets}. However, since the shapes of the primitives are constrained to be convex, the interpretability of shapes is limited to part parsing. In this work, we study the star domain as a primitive representation that has more expressive power than that of previously proposed primitive representations.

In computation theory, 2D polygonal shape decomposition by star domain has a long history of work \cite{chvatal1975combinatorial, keil1985decomposing}. In computer vision, star domain has been used to abstract 3D shapes to \emph{encode} shape embedding \cite{liu2003directional, poulenard2019effective, cohen2018spherical, kondor2018clebsch} for discriminative models. In contrast, we study the application of star domain to \emph{decode} shape embedding to accurately reconstruct 3D shapes for generative models.

Surface representation of 3D objects in the context of generative models has been studied extensively. In recent studies, the standard explicit shape representation for generative models is a mesh \cite{groueix2018papier, kanazawa2018end, ranjan2018generating, wang2018pixel2mesh, groueix2018papier}. Mesh \cite{gao2019sdm}, pointcloud \cite{deprelle2019learning}, and parametrized surface \cite{tulsiani2017learning, niu2018im2struct, Paschalidou2019CVPR, Paschalidou2020CVPR} have been studied as explicit surfaces for primitive models. A state-of-the-art method employs a learnable indicator function for non-primitive \cite{mescheder2019occupancy, park2019deepsdf} and primitive based approaches \cite{genova2019learning, chen2019bsp, deng2019cvxnets}. However, extracting a surface mesh during inference is quite costly, as the isosurface extraction operation grows cubically for desired meshing resolutions. An implicit representation model with fast polymesh sampling during inference has been proposed in \cite{chen2019bsp}. However, due to the lack of explicit surface representation during training, primitive based methods with implicit representations need complicated training schemes such as near surface training data sampling with ray casting \cite{genova2019learning, deng2019cvxnets}, and heuristic losses to keep primitives inside the shape boundary \cite{deng2019cvxnets}, or a multi-stage training strategy to approximate explicit surfaces \cite{chen2019bsp}. Notable exception is \cite{liao2018deep} which has both implicit and explicit representaions, however, it is possible by reconstructing the shape as voxel at the cost of limited shape resolution. In this study, we propose unified shape representation in both explicit and implicit forms in arbitrary resolution. Our approach utilizes this advantage to realize a simple training scheme with fast high-resolution mesh sampling during inference.

\section{Methods}
We begin by formulating the problem setting in Section \ref{sec:setting}. Next, we define star domain in Section \ref{sec:star}. Also, we introduce NSD to approximate shapes in the star domain with a theoretical analysis of the representation power. Using NSD as a building block, we describe the pipeline of our approach in Section \ref{sec:NSDdecoder}, \ref{sec:NSDN} and \ref{sec:loss}. Implementation details are provided in Section \ref{sec:impldetail}.

\subsection{Problem setting}
\label{sec:setting}
We represent an object shape as set of surface points $P\subseteq \mathbb{R}^3$, and as an indicator function which can be evaluated at an arbitrary point ${\bf x} \in \mathbb{R}^3$ in 3D space as $O: \mathbb{R}^3 \rightarrow \{0, 1\}$, where $\{{\bf x}\in \mathbb{R}^3 \,|\, O(x) = \tau\}$. In this equation, $\tau = 0$ defines the outside of the object, $\tau = 1$ defines the inside. Our goal is to parametrize the 3D shape by a composite indicator function $\hat O$, and surface points $\hat P$ which can be decomposed into a collection of $N$ primitives. The $i$ th primitive has an indicator function ${\hat O_i}: \mathbb{R}^3 \rightarrow [0, 1]$, and a surface point function defined on a sphere ${\hat P_i}: \mathbb{S}^2 \rightarrow \mathbb{R}^3$. To realize implicit and explicit shape representation simultaneously, we further require ${\hat O}$ and ${\hat P}$ to be related as ${\hat O}({\hat p}) = \tau_o$, where ${\hat p} \in {\hat P}$ and $\tau_o \in [0, 1]$ is a constant to represent isosurface. We ensure that both the composite indicator function and the surface points are approximated as ${O} \approx {\hat O}$ and ${P} \approx {\hat P}$, respectively through training losses.
 
\subsection{Neural star domain}
\label{sec:star}
A geometry $U \subseteq \mathbb{R}^3$ is a star domain if $\exists {\bf t} \in U, \forall {\bf u} \in U, [{\bf t}, {\bf u}] = \{ (1-v){\bf t} + v{\bf u}, 0 \leq v \leq 1 \} \subseteq U$. Intuitively, any geometry which has an origin ${\bf t}$, such that a straight line segment between any point ${\bf u}$ inside the geometry and ${\bf t}$ does not have self-intersection, belongs to the star domain. Thus, we can regard the star domain shapes as continuous functions defined on the surface of a sphere. We denote such functions as $r: {\mathbb{S}^2 \rightarrow \mathbb{R}}$. The spherical harmonics expansion $\mathbb{S}^2 \rightarrow \mathbb{R}$ is a multivariate polynomial function which is also defined on the surface of a sphere. Thus, we can formulate the star domain using a spherical harmonics expansion as:
\begin{equation}
\label{eq:sh}
r({\bf d}) = \sum_{l=0}^{\infty}\sum_{m=-l}^{l} c_{l,m} Y_{l,m}(\omega({\bf d})),\ \omega({\bf d})=(\sin \theta \cos \phi, \sin \theta \sin \phi, \cos \theta)
\end{equation}
where ${\bf d} = (\theta, \phi)\in \mathbb{S}^2$, $c_{l, m}\in \mathbb{R}$ is a constant, and $Y_{l, m}$ is the Cartesian spherical harmonic function \cite{varshalovich1988quantum}. A Few examples of $Y_{l, m}$ given $(l, m)$ can be found in the Appendix. 

In order to realize the star domain primitive, we propose Neural Star Domain (NSD), which approximates $r$ by a neural network $f_{NN}$ taking $\omega(\cdot)$ as input. 

\paragraph{Approximation ability.}
\label{sec:proof}
We demonstrate the universal approximation ability of NSD to a star domain $r$.
Our theorem says that $r$ can be arbitrarily approximated by NSD.
\begin{theorem*}
\label{th:theorem}
  Suppose $r: \mathbb{S}^2 \rightarrow \mathbb{R}$ is a continuous function on the surface of a sphere, for $\forall \epsilon > 0$, $\exists$ a neural star domain $f_{NN}\circ \omega:\mathbb{S}^2 \rightarrow \mathbb{R}$, such that for any ${\bf d} \in \mathbb{S}^2$,
  \begin{equation}
    \left| r({\bf d}) - f_{NN}(\omega({\bf d})) \right| < \epsilon
  \end{equation}
\end{theorem*}

\begin{proof}
By the completeness of spherical harmonics \cite{burkhardt2008foundations} to a continuous function on a spherical surface as shown in Equation \eqref{eq:sh}, $\forall \epsilon_1 > 0$, $\exists L\in \mathbb{N}^+$, $\exists c_{l,m} \in \mathbb{R}^+$, for any ${\bf d} \in \mathbb{S}^2$:
\begin{equation}
\label{eq:bound1}
\left| r({\bf d}) - r_L({\bf d}) \right| < \epsilon_1, \ {\rm where}\  r_L({\bf d}) = \sum_{l=0}^{L}\sum_{m=-l}^{l}c_{l,m}Y_{l,m}(\omega({\bf d}))
\end{equation}
$\omega$ can be regarded as $Y_{1, m}$ with a proper constant $c_{1, m}$, and from the definition of Cartesian spherical harmonics \cite{varshalovich1988quantum}, each $Y_{l, m}$ with $l > 1$ can be written as a polynomial function of $Y_{1, m}$ with a proper constant $c_{l, m}$. Thus, $r_L$ can be regarded as a polynomial function over $\omega$, i.e., it is continuous over $\omega$.

Now, by the universal approximation theorem of neural networks to a continuous function \cite{hornik1991approximation, arora2016understanding}, $\forall \epsilon_2 > 0$, $\exists$ a neural network $f_{NN}:\mathbb{R}^3 \rightarrow \mathbb{R}$, such that for any $\omega_{{\bf d}} \in \{\omega({\bf d})\,|\,{\bf d}\in\mathbb{S}^2\}$,
\begin{equation}
\label{eq:bound2}
\left|r_L({\bf d})- f_{NN}(\omega_{{\bf d}}) \right| < \epsilon_2
  \end{equation}
Given Equations \eqref{eq:bound1} and \eqref{eq:bound2}, $\forall \epsilon > 0$, $\exists$ a neural network $f_{NN}:\mathbb{R}^3 \rightarrow \mathbb{R}$, such that for any ${\bf d} \in \mathbb{S}^2$:
\begin{equation}
    \left| r({\bf d}) - f_{NN}(\omega({\bf d})) \right| < \epsilon_1 + \epsilon_2 = \epsilon
\end{equation}
\end{proof}

Note that network architectures that take output values of trigonometric functions as input exist, such as HoloGAN \cite{nguyen2019hologan}. However, our approach differs in input and output as follows: (1) By taking $\omega$ as an input, our approach is theoretically grounded in approximate spherical harmonic expansion. HoloGAN takes output values of high-degree trigonometric polynomial functions as input. (2) The neural network in HoloGAN targets to predict high-dimensional vectors as images, whereas ours specifically targets predicting a single-dimensional radius r by approximating star domain.
 
 \subsection{Primitive representation}
 \label{sec:NSDdecoder}
 As NSD is defined on the surface of a sphere, one can define both implicit and explicit shape representations of a primitive. For simplicity, we define NSD $f:=f_{NN} \circ \omega$ in following sections.
 
\paragraph{Implicit representation.}
Given the 3D location ${\bf x}\in \mathbb{R}^3$, an indicator function ${\hat O_i}:{\mathbb R}^3 \rightarrow [0, 1]$ for the $i$ th primitive located at ${\bf t}_i$ is expressed as follows:
\begin{equation}
\label{eq:pr_impl}
  {\hat O_i}({\bf x};{\bf t}_i) = {\rm Sigmoid}(\alpha (1 - \frac{\lVert {\bf \bar {x}} \rVert}{r^+})),\ {\rm where}\ {\bf {\bar x}} = {\bf x} - {\bf t}_i,\ r^+ = {\rm ReLU}(f_i(G({\bf x})))
\end{equation}
where $\alpha$ is the scaling factor that adjusts the margin of the indicator values between the inside and outside of the shape, $G: {\mathbb R}^3 \rightarrow \mathbb{S}^2$ denotes conversion from 3D Cartesian coordinates to the spherical surface, and the ReLU operator ensures that the estimated radius is a non-negative real value. Note that $\lVert {\bf \bar {x}} \rVert - r^+$ can be interpreted as the singed distance function. The formulae of $G$ and $G^{-1}$ can be found in the Appendix.

\begin{figure}
\centering
      \includegraphics[width=14cm]{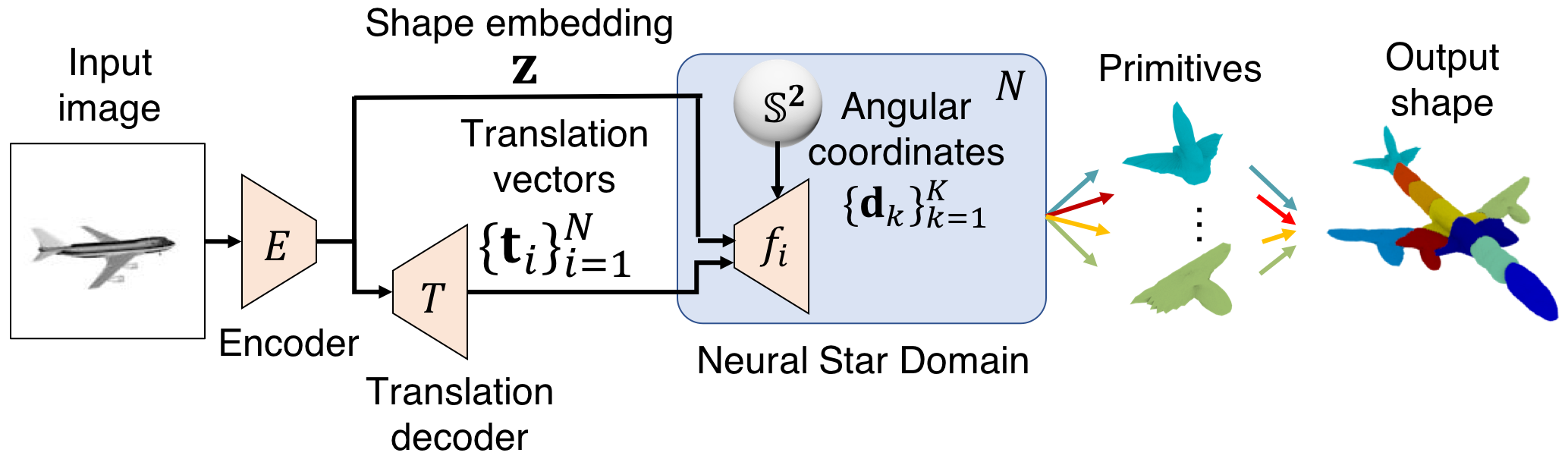}
            \caption{Architecture of NSDN.}
            \label{fig:pipeline}
\end{figure}
\vspace{-1\baselineskip}

\paragraph{Explicit representation.}
With a slight abuse of notation, we denote a conversion from spherical coordinates to the 3D location as $G^{-1}: {\mathbb R} \times {\mathbb S}^2 \rightarrow {\mathbb R}^3$. We can sample a surface point in the direction of ${\bf d}$ from the origin of the $i$ th primitive located at ${\bf t}_i$ as follows:
\begin{equation}
  {\hat P_i}({\bf d};{\bf t}_i) = G^{-1}(r^+, {\bf d}) + {\bf t}_i,\ {\rm where}\ r^+ = {\rm ReLU}(f_i({\bf d}))
\end{equation}

\subsection{Neural star domain network (NSDN)}
\label{sec:NSDN}
To represent the target shape as a collection of primitives, we define an NSDN. NSDN employs the bottleneck auto-encoder architecture, similar to \cite{mescheder2019occupancy}. NSDN consists of an encoder $E$, a translation network $T$, and a set of NSDs $\{f_i\}^N_{i=1}$. Given an input $I$, the encoder $E$ derives a shape embedding ${\bf z}$. Then, the translation network $T$ outputs a set of translation vectors $\{{\bf t}_i\}^N_{i=1}$ from ${\bf z}$. Translation vectors represent the locations of each primitive. The $i$ th NSD $f_i$ acts as a decoder, and infers the radius given an angular coordinate ${\bf d}$, translation vectors ${\bf t}_i$ and a shape embedding ${\bf z}$. In this study, we only estimate the location as pose of the primitives, whereas previous works additionally predicting the scale and the rotation of each primitive \cite{deng2019cvxnets, Paschalidou2019CVPR, tulsiani2017learning}. We observe that learning rotation and scale leads to unsuccessful training. The overview of the architecture is illustrated in Figure \ref{fig:pipeline}.

\paragraph{Composite indicator function.}
To derive an implicit representation of NSDN, we define a composite indicator function as the union of $N$ NSD indicator functions as:
\begin{equation}
  {\hat O}({\bf x}; \{{\bf t}_i\}^N_{i=1}) = {\rm Sigmoid}({\sum_{i \in [N]}{{\hat O_i}({\bf x};{\bf t}_i)}})
\end{equation}
To encourage gradient learning of all primitives during training, we take the sum of the indicator values over the primitives rather than the maximum value. We treat the threshold of the indicator value $\tau_o$ of the surface level of ${\hat O}$ as a hyperparameter.

\paragraph{Surface point extraction.}
Owing to the unified explicit and implicit shape representation of NSD, NSDN is able to extract the union of surface points of primitives in a differentiable manner. We define the unified surface points as follows:
\begin{equation}
    \label{eq:surfacepoint}
{\hat P} = \bigcup_{i}{\{{\hat P}_i({\bf d}; {\bf t}_i)|\, \forall j \in [N\setminus i],\,  {\hat O_j}({\hat P}_i({\bf d}; {\bf t}_i); {\bf t}_i)<\tau_s,\, {\bf d} \in \{{\bf d}_k\}^K_{k=1}\}}
\end{equation}

where $K$ denotes the number of points sampled from the surface of the sphere, and $\tau_s$ is a hyperparameter for the threshold of the indicator value for the surface points.

\subsection{Training loss}
\label{sec:loss}
To learn the parameters ${\bf \Theta}$ of NSDN, we define the \emph{surface point loss} that minimizes the symmetric Chamfer distance between the surface points $P$ from a training sample, and those from predicted surface points ${\hat P}$. The surface point loss is formulated as:
\begin{equation}
    \label{eq:surfacepointloss}
  L_S({\bf \Theta}) = \mathbb{E}_{{\hat p}\sim {\hat P}}\min_{p\sim P}\lVert {\hat p}-p \rVert + \mathbb{E}_{p\sim P}\min_{{\hat p}\sim {\hat P}}\lVert p - {\hat p} \rVert
\end{equation}
Note that surface point loss enables learning collective surfaces of primitives by accessing both implicit and explicit representations as shown in Equation \ref{eq:surfacepoint}. The training loss leads to a better reconstruction than minimizing the distance between $P$ and a simple union of the surface points of the primitives $\bigcup_{i\in [N]}{\{{\hat P}_i({\bf d}; {\bf t}_i)| {\bf d} \in \{{\bf d}_k\}^K_{k=1}\}}$ as in \cite{Paschalidou2019CVPR, tulsiani2017learning}. This is because, ideally, the loss should measure the distance between the two sets of surface points. We also use the occupancy loss as in \cite{mescheder2019occupancy} $L_O({\bf \Theta}) = \mathbb{E}_{{\bf x}\sim\mathbb{R}^3}{\rm BCE}(O({\bf x}), {\hat O}({\bf x}))$, where ${\rm BCE}$ is the binary cross entropy. We observe that using the occupancy loss in addition to the surface point loss achieves the best reconstruction performance.

\subsection{Implementation details}
\label{sec:impldetail}
In all of our experiments, we use the same architecture while varying the number of primitives $N$. $N$ is set to 30 by default, unless stated otherwise. We use ResNet18 as the encoder $E$ that produces shape embedding as a latent vector ${\bf z}\in \mathbb{R}^{256}$ for an input RGB image by following OccNet \cite{mescheder2019occupancy}. For the translation network $T$, we use a multilayer perceptron (MLP) with three hidden layers with $(128, 128, N * 3)$ units with ReLU activation. For NSD, we use a MLP with three hidden layers with $(64, 64, 1)$ units with ReLU activation. We use 100 for the margin $\alpha$ of the indicator function. The threshold $\tau_o$ of the composite indicator function is determined by a grid search over the validation set. For example, for $N=30$, we use $\tau_o=0.99$. We use 0.1 for the threshold $\tau_s$ of surface point extraction. During training, we use a batch size of 20, and train with the Adam optimizer with a learning rate of 0.0001. We set the weight of $L_o$ and $L_s$ as 1 and 10, respectively. For the training data, we sample 4096 points from the ground truth pointcloud, and $400 * N$ samples from the generated shape for the surface point loss $L_s$, and sample 2048 points from the ground truth indicator values for the indicator loss $L_o$. For mesh sampling, we use a spherical mesh template.

\section{Experiments}
\paragraph{Dataset.}
In our experiments, we used the ShapeNet \cite{chang2015shapenet} dataset. By following \cite{mescheder2019occupancy}, we test our approach on thirteen categories of objects. In addition, we use the same samples and data split as those in \cite{mescheder2019occupancy}. For 2D images, we use the rendered view provided by \cite{choy20163d}. For the quantitative evaluation of the part semantic segmentation, we use PartNet \cite{Mo_2019_CVPR} and part labels provided by \cite{chen2019bae}.

\paragraph{Methods.}
We compare our approach against several state-of-the-art approaches with different shape representations. As primitive based reconstruction approach, we compare against BSP-Net \cite{chen2019bsp}, CvxNet \cite{deng2019cvxnets}, and SIF \cite{genova2019learning} as the implicit representation based approaches, and AtlasNetV2 \cite{deprelle2019learning} for the explicit representation based approach. Since \cite{chen2019bsp, deng2019cvxnets} represent shapes as a collection of convexes, we regard them as a baseline for the effectiveness of the star domain primitive representation. As a non-primitive based reconstruction approach, we compare against OccNet \cite{mescheder2019occupancy} as the leading implicit representation based technique, and AtlasNet \cite{groueix2018papier} for the explicit shape representation. For AtlasNetV2, since the code provided by the author does not include a model for single view reconstruction, we replace the provided encoder with the same ResNet18 used by NSDN and OccNet, and train the model from scratch. Furthermore, to make a fair comparison with NSDN, we sample 400 points from each patch during training, and use 30 patches for AtlasNetV2, unless otherwise noted. We confirm that this leads to a slightly better reconstruction accuracy than the original configuration. For BSP-Net, we use the pretrained model described in Section \ref{sec:svr_recon}. In Section \ref{semantic}, we train BSP-Net from scratch with the provided code of \cite{chen2019bsp}. Since BSP-Net uses different train and test splits, we evaluated it on the intersection of the test splits from \cite{mescheder2019occupancy} and \cite{chen2019bsp}. 

\paragraph{Metrics.}
We evaluate our methods with reconstruction accuracy, part correspondence and mesh sampling speed.
For evaluation on the reconstruction accuracy, we apply three commonly used metrics to compute the difference between the meshes of reconstruction and groud truth: (1) F-score; by following the argument of \cite{TB19}, it can be interpreted as the percentage of correctly reconstructed surfaces. (2) L1 Chamfer distance (CD1), and (3) volumetric IoU (IoU). For all metrics, we use 100,000 sample points from the ground truth meshes, and reconstruct shape meshes by following \cite{mescheder2019occupancy, deng2019cvxnets}. For evaluation of the part correspondence in semantic capability, we use the standard label IoU between the ground truth part label and the predicted part label. For mesh sampling speed, we measure the time in which a pipeline encodes an image and decodes the mesh vertices and faces. We exclude the time for device I/O. we conduct all speed measurements on an NVIDIA V100 GPU. Moreover, for fair comparison, we measure the time to mesh a single primitive for AtlasNet, AtlasNetV2, and BSP-Net as an analogy of parallel processing, because their original implementation sequentially process each primitive for meshing, whereas our implementation does so in parallel.

\begin{table}
  \centering
  \resizebox{\textwidth}{!}{
\setlength{\tabcolsep}{3pt}

  \begin{tabular}{c|c|ccccccccccccc|c|c}
      \bhline{1.5 pt}


 &  & airplane & bench & cabinet & car & chair & display & lamp & speaker & rifle & sofa & table & phone & vessel & mean & time \\ \hline
\multirow{8}{*}{F-score} & AtlasNet \cite{groueix2018papier} & 67.24 & 54.50 & 46.43 & 51.51 & 38.89 & {\bf 42.79} & 33.04 & 35.75 & {\bf 64.22} & 43.46 & 44.93 & 58.85 & {\bf 49.87} & 48.57 &  0.008\\ 
 & AtlasNetV2  \cite{deprelle2019learning} & 54.99 & 50.67 & 31.95 & 39.73 & 29.10 & 33.55 & 28.35 & 22.54 & 62.27 & 30.15 & 45.93 & 51.45 & 39.91 & 40.05&  0.010\\ 
 & OccNet \cite{mescheder2019occupancy} & 62.87 & 56.91 & 61.79 & 56.91 & 42.41 & 38.96 & 38.35 & 42.48 & 56.52 & 48.62 & 58.49 & 66.09 & 42.37 & 51.75 & 0.525\\ 
 & OccNet* \cite{mescheder2019occupancy}  & 63.56 & 57.39 & {\bf 63.03} & 61.41 & {\bf 43.61} & 41.54 & {\bf 41.13} & {\bf 45.39} & 57.94 & {\bf 49.86} & {\bf 59.62} & 66.11 & 45.00 & 53.51 & 0.529\\ 
 & SIF \cite{genova2019learning} & 52.81 & 37.31 & 31.68 & 37.66 & 26.90 & 27.22 & 20.59 & 22.42 & 53.20 & 30.94 & 30.78 & 45.61 & 36.04 & 34.86 & n/a\\ 
 & CvxNet \cite{deng2019cvxnets}& {\bf 68.16} & 54.64 & 46.09 & 47.33 & 38.49 & 40.69 & 31.41 & 29.45 & 63.74 & 42.11 & 48.10 & 59.64 & 45.88 & 47.36& n/a\\ 
 & BSP-Net \cite{chen2019bsp} & 61.91 & 53.12 & 44.75 & 55.24 & 38.57 & 35.68 & 29.98 & 34.04 & 57.28 & 43.89 & 46.42 & 49.18 & 42.76 & 45.60 & 0.014\\ 
 & NSDN (ours)& 67.96 & {\bf 60.37} & 59.26 & {\bf 63.54} & 43.58 & 41.81 & 38.83 & 43.09 & 63.31 & 48.97 & 57.91 & {\bf 70.65} & 46.49 & {\bf 54.29} &0.014 \\ \hline 
 
\multirow{9}{*}{CD1} & AtlasNet  \cite{groueix2018papier} & 0.104 & 0.138 & 0.175 & 0.141 & 0.209 & {\bf 0.198} & {\bf 0.305} & {\bf 0.245} & 0.115 & 0.177 & 0.190 & 0.128 & {\bf 0.151} & 0.175 & 0.008 \\ 
 & AtlasNetV2  \cite{deprelle2019learning} & 0.119 & 0.164 & 0.246 & 0.176 & 0.256 & 0.209 & 0.313 & 0.340 & {\bf 0.099} & 0.210 & 0.221 & 0.131 & 0.159 & 0.203  & 0.010 \\ 
 & OccNet \cite{mescheder2019occupancy}  & 0.147 & 0.155 & 0.167 & 0.159 & 0.228 & 0.278 & 0.479 & 0.300 & 0.141 & 0.194 & 0.189 & 0.140 & 0.218 & 0.215 & 0.525  \\ 
 & OccNet*  \cite{mescheder2019occupancy} & 0.141 & 0.154 & {\bf 0.149} & 0.150 & 0.206 & 0.214 & 0.369 & 0.254 & 0.142 & 0.182 & 0.175 & 0.124 & 0.194 & 0.189 & 0.529 \\ 
 & SIF \cite{genova2019learning}& 0.167 & 0.261 & 0.233 & 0.161 & 0.380 & 0.401 & 1.096 & 0.554 & 0.193 & 0.272 & 0.454 & 0.159 & 0.208 & 0.349& n/a \\ 
 & CvxNet  \cite{deng2019cvxnets} & {\bf 0.093} & {\bf 0.133} & 0.160 & {\bf 0.103} & 0.337 & 0.223 & 0.795 & 0.462 & 0.106 & {\bf 0.164} & 0.358 & {\bf 0.083} & 0.173 & 0.245 & n/a  \\ 
 & BSP-Net  \cite{chen2019bsp} & 0.128 & 0.158 & 0.179 & 0.153 & 0.211 & 0.224 & 0.332 & 0.269 & 0.126 & 0.190 & 0.190 & 0.153 & 0.189 & 0.192 & 0.014  \\ 
 & NSDN  (ours)& 0.111 & 0.135 & 0.155 & 0.136 & {\bf 0.191} & 0.205 & 0.320 & 0.251 & 0.118 & 0.177 & {\bf 0.167} & 0.110 & 0.174 & {\bf 0.173} & 0.014 \\ \hline

\multirow{7}{*}{IoU} & OccNet  \cite{mescheder2019occupancy}  & 0.571 & 0.485 & 0.733 & 0.737 & 0.501 & 0.471 & 0.371 & 0.647 & 0.474 & 0.680 & 0.506 & 0.720 & 0.530 & 0.571& 0.525\\ 
 & OccNet*  \cite{mescheder2019occupancy} & 0.591 & {\bf 0.492} & {\bf 0.750} & {\bf 0.746} & {\bf 0.530} & 0.518 & {\bf 0.400} & {\bf 0.677} & 0.480 & {\bf 0.693} & {\bf 0.542} & 0.746 & 0.547 & {\bf 0.593}& 0.529 \\ 
 & SIF \cite{genova2019learning} & 0.530 & 0.333 & 0.648 & 0.657 & 0.389 & 0.491 & 0.260 & 0.577 & 0.463 & 0.606 & 0.372 & 0.658 & 0.502 & 0.499 & n/a\\ 
 & CvxNet  \cite{deng2019cvxnets}& 0.598 & 0.461 & 0.709 & 0.675 & 0.491 & {\bf 0.576} & 0.311 & 0.620 & 0.515 & 0.677 & 0.473 & 0.719 & {\bf 0.552} & 0.567 & n/a\\ 
 & BSP-Net  \cite{chen2019bsp} & 0.549 & 0.371 & 0.660 & 0.708 & 0.466 & 0.507 & 0.323 & 0.638 & 0.462 & 0.667 & 0.428 & 0.711 & 0.523 & 0.539 & 0.014\\ 
 & NSDN  (ours) & {\bf 0.613} & 0.461 & 0.719 & 0.742 & 0.515 & 0.553 & 0.368 & 0.667 & {\bf 0.516} & 0.689 & 0.511 & {\bf 0.760} & 0.550 & 0.589& 0.014 \\


\bhline{1.5 pt}
\end{tabular}
}
        \vspace{+0.5\baselineskip}
\caption{
Reconstruction performance on ShapeNet \cite{chang2015shapenet}. In the far right column of the table, denoted as time, we report per object average duration (in seconds) of mesh sampling to show the time cost to produce an evaluated mesh. Since we do not perform data augmentation as opposed to the original implementation of OccNet \cite{mescheder2019occupancy}, we also report the results of pretrained OccNet trained without data augmentation denoted as OccNet*.}
\label{tb:svrresults}
        \vspace{-1\baselineskip}
\end{table}

\begin{figure}
\vspace{-1\baselineskip}
\begin{tabular}{@{}c@{}}

\begin{minipage}{0.48\textwidth}
\begin{center}
      \includegraphics[width=6.5cm]{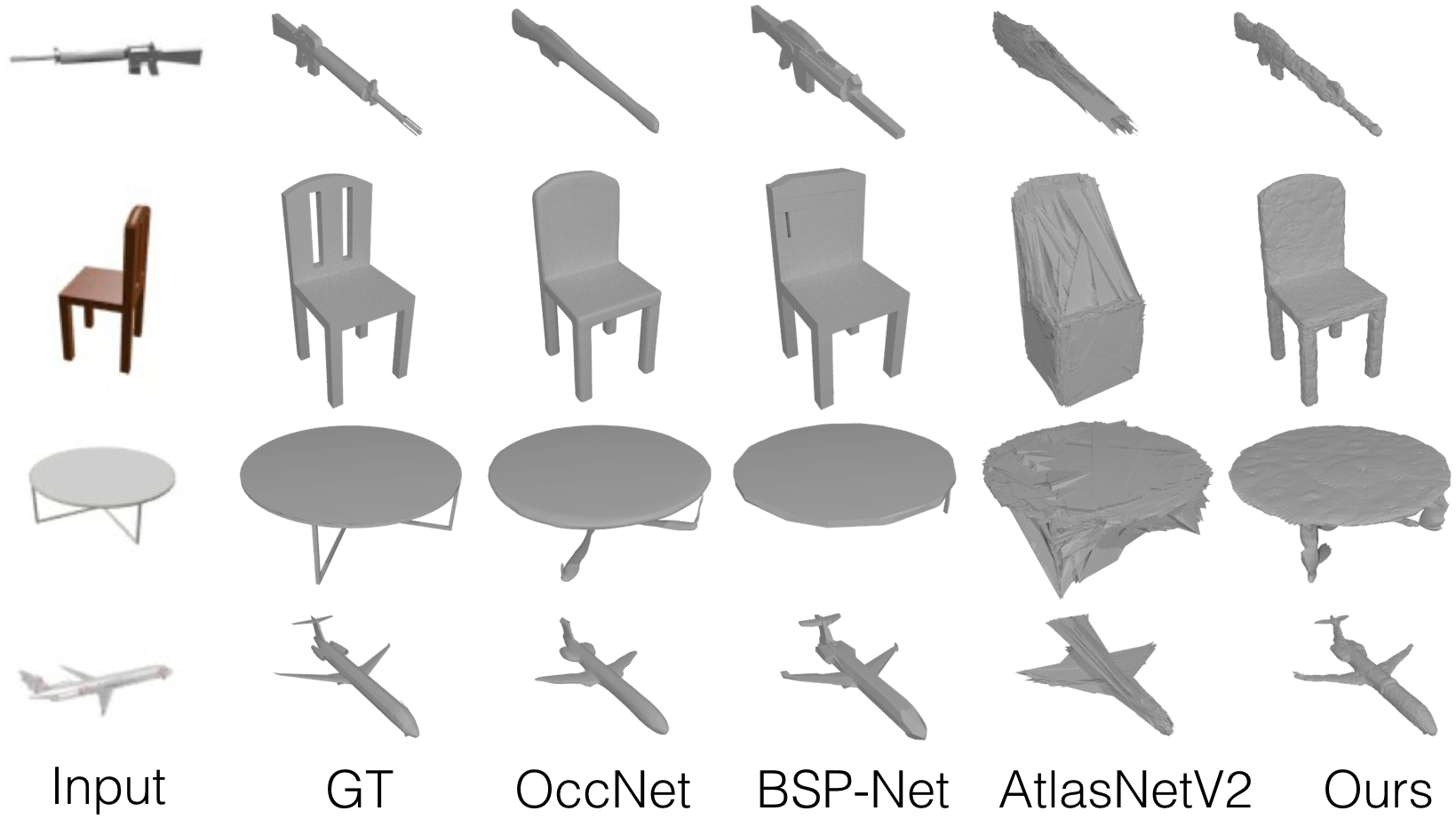}
        \caption{Visualization of reconstructed meshes with an RGB image input. Best viewed zoomed in.}

\label{fig:svrimages}
\end{center}

\end{minipage}

\hspace{0.01\textwidth}

\begin{minipage}{0.5\textwidth}
\begin{center}
\makeatletter
\def\@captype{table}
\makeatother


\begin{tabular}{c|c@{\hspace*{2mm}}c|c}
\bhline{1.5 pt}

&implicit&explicit&F-score \\ \hline 
AtlasNetV2 \cite{deprelle2019learning}&  & \checkmark & 40.05 \\ 
BSP-Net \cite{chen2019bsp}& \checkmark &  & 45.60 \\ 
NSDN${}_O$ & \checkmark &  & 23.93 \\ 
NSDN${}_C$ &  & \checkmark & 45.84 \\ 
NSDN${}_S$ & \checkmark & \checkmark & 50.52 \\ 
NSDN${}_{S+O}$ & \checkmark & \checkmark & {\bf 52.27} \\ 

\bhline{1.5 pt}
\end{tabular}
\caption{Effects of different losses on the F-score. In the table, check marks under the implicit and explicit columns denote if the loss uses a corresponding shape representation. In NSDN, $O$ denotes using only occupancy loss, $C$ for using only Chamfer loss without surface point extraction, and $S$ for using only surface point loss.}
\label{tb:effect_of_loss}

\end{center}
\end{minipage}

\end{tabular}
\vspace{-1\baselineskip}

\end{figure}

\subsection{Single view reconstruction}
\label{sec:svr_recon}
We evaluate the reconstruction performance of NSD against state-of-the-art methods for an input RGB image. The quantitative result is shown in Table \ref{tb:svrresults}. Qualitative examples are shown in Figure \ref{fig:svrimages}. The number of faces of the generated meshes of NSDN and AtlasNetV2 are made comparable with those of OccNet \cite{mescheder2019occupancy}. We find that: (1) NSDN consistently outperforms previous primitive based approaches (CvxNet, SIF, BSP-Net, AtlasNetV2) in terms of the averages of all metrics. Significant improvement is seen especially in terms of the F-score. (2) NSDN works relatively better than the leading technique (OccNet \cite{mescheder2019occupancy}) in the CD1 and F-score. Note that our method performs comparably with OccNet, but the mesh sampling speed is distinctively faster. Details of the mesh sampling analysis can be found in Subsection \ref{sec:mesh_gen}.

\paragraph{Effect of losses.}
As the surface point loss is made available by using integrated implicit and explicit representation, we evaluate the effectiveness of the proposed loss to show the unique benefit of our representation. We use $N=10$ for faster NSDN training to accelerate experiments. The result is shown in Table \ref{tb:effect_of_loss}. Using only occupancy loss leads to unsuccessful training. Using the standard Chamfer loss leads to comparable performance with previous works. Using the surface point loss outperforms leading primitive-based techniques \cite{chen2019bsp}. Additionally, using occupancy loss along with surface point loss leads to a slightly higher accuracies and achieves the best results.

\subsection{Semantic capability}
\label{semantic}
We evaluate the semantic capability of our approach against other approaches with implicit and explicit primitive representations: BSP-Net \cite{chen2019bsp} and AtlasNetV2 \cite{deprelle2019learning}. Following the evaluation methods in \cite{deng2019cvxnets, chen2019bsp, Paschalidou2019CVPR}, with varying number of primitives for each method, we evaluate the semantic capabilities of the approaches as a trade-off between representation parsimony and (1) reconstruction accuracy measured by the F-score. For the semantic segmentation task, labels for each ground truth point are predicted as follows, and (2) semantic segmentation accuracy on part labels. (1) For each ground truth point in a training sample, find the nearest primitive to it and vote for the part label of the point, (2) assign each primitive a part label that has the highest number of votes, and (3) for each point of a test sample, find the nearest primitive and assign the part label of the primitive to that point. We use four classes for semantic segmentation: plane, chair, table, and lamp. For the table and lamp, we follow \cite{chen2019bsp} to reduce the parts from (base, pole, lampshade, canopy) $\rightarrow$ (base, pole, lampshade), and analogously for table (top, leg, support) $\rightarrow$ (top, leg). The models are trained without any part label supervision. 

\begin{figure}

  \centering
      \includegraphics[width=13.5cm]{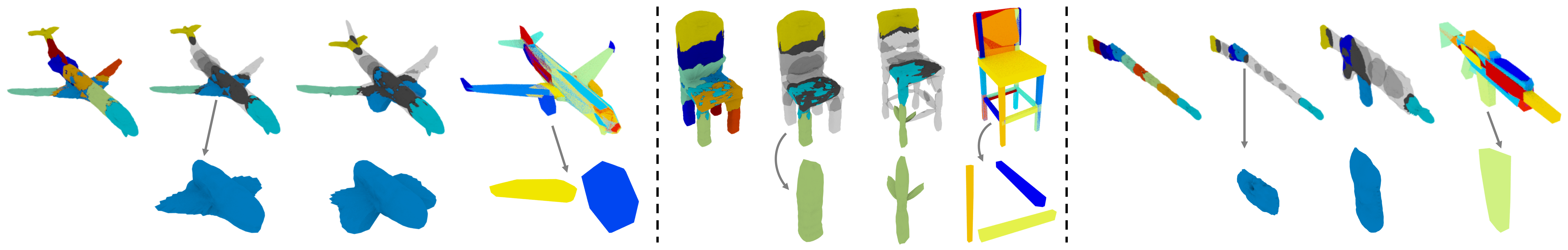} 
        \caption{Visualization of the primitives of different categories (plane, chair, and rifle). In each category, from far left: (1) reconstruction results with colored primitives, (2) top: only a few primitives are colored to show part correspondence with another reconstruction result on the right. Bottom: one primitive is chosen and zoomed. (3) Top: another reconstruction result in the same category. Bottom: Same primitive as in the previous visualization. (4) Top: Reconstruction result of same object with previous reconstruction by BSP-Net. Bottom: Manually selected primitives that correspond to the same semantic parts of the previous primitives. Best viewed zoomed in color.}
\label{fig:primitive_visualize}
\end{figure}

\begin{figure}
\vspace{-1\baselineskip}

\begin{tabular}{@{}c@{}}

\begin{minipage}{0.45\textwidth}
\begin{center}

      \includegraphics[width=6.0cm]{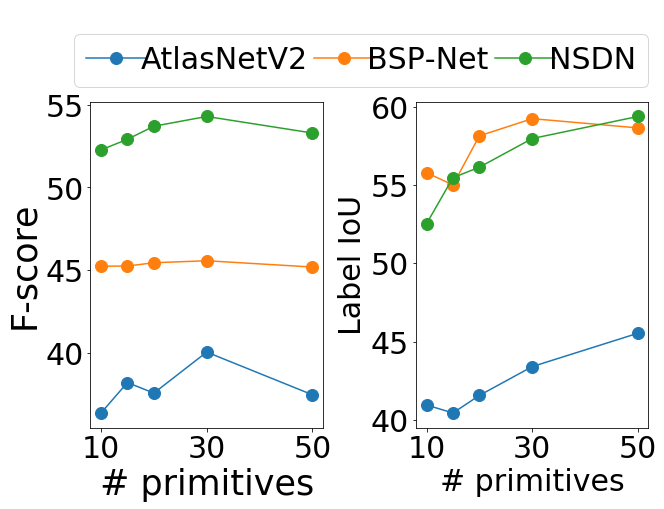}

\vspace{-0.5\baselineskip}
\caption{F-score and label IoU with variying number of primitives. Number of primitives evaluated are: 10, 15, 20, 30, 50.}
\label{fig:varying_parts_vis}

\end{center}
\end{minipage}

\hspace{0.01\textwidth}

\begin{minipage}{0.52\textwidth}
\begin{center}

\makeatletter
\def\@captype{table}
\makeatother

\begin{tabular}{ccccc}
\bhline{1.5 pt}

&$\#$V& $\#$F &F-score&time\\ \hline 
NSDN ico0 & 2 & 5 & 34.02 & 0.012 \\ 
NSDN ico2 & 30 & 88 & 42.87 & 0.013 \\ 
NSDN ico4 & 478 & 1414 & 55.66 & 0.017 \\ 
MISE up0 & 12 & 31 & 26.46 & 0.051 \\ 
MISE up1 & 54 & 143 & 40.37 & 0.635 \\ 
MISE up2 & 220 & 592 & 50.28 & 5.438 \\ 
BSP-Net \cite{chen2019bsp}& 
\textcolor[gray]{0.5}{ 10 }&
\textcolor[gray]{0.5}{ 18 }&
 45.60 &
 0.014\\


\bhline{1.5 pt}
\end{tabular}

\caption{Mesh sampling speed for given mesh properties. $\#V$ and $\#F$ denote the numbers of mesh vertices ($\times 100$) and mesh faces ($\times 100$), respectively. Ico$\#$ denotes number of icosphere subdivisions used as the mesh template of the primitive. Up$\#$ denotes the number of upsampling steps in MISE \cite{mescheder2019occupancy}. Up0 equals to $32^3$ voxel sampling and up2 to $128^3$.}

\label{tb:meshtable}

\end{center}

\end{minipage}

\end{tabular}

\end{figure}

In Figure \ref{fig:varying_parts_vis}, we show that our method consistently outperforms previous methods in reconstruction accuracy regardless of the number of the primitives, while performing comparatively in the semantic segmentation task. This demonstrates the superior semantic capabilities of our approach. While maintaining comparable performance in consistent part correspondence with the previous work \cite{chen2019bsp}, our approach better reconstructs target shapes. We visualize the learned primitives in Figure \ref{fig:primitive_visualize}. We can see that our approach is more parsimonious in reconstructing corresponding parts of \cite{chen2019bsp}.

\paragraph{Effect of overlap regularization.}
High expressivity of NSD results in a severe overlap of the primitives, which leads to less interpretability of part correspondence. To alleviate this problem, we investigate the effect of the overlapping regularization.
By exploiting NSD is also implicit representation, we adapt the decomposition loss proposed in \cite{deng2019cvxnets} as an off-the-shelf overlap regularizer. The hyperparameter $\tau_r$ controls the amount of the overlap. The definition of the regularizer can be found in the appendix. We set the loss weight of the regularizer to 10. In this experiment, we train the model for the airplane and chair categories. As we found out that the optimal $\tau_r$ varies across categories, we train our model with a single category. We use 1 and 1.2 for $\tau_r$ in the airplane and the chair categories, respectively.

\begin{figure}
\vspace{-0.5\baselineskip}

\begin{tabular}{@{}c@{}}

\begin{minipage}{0.51\textwidth}
\begin{center}

\includegraphics[width=7.0cm]{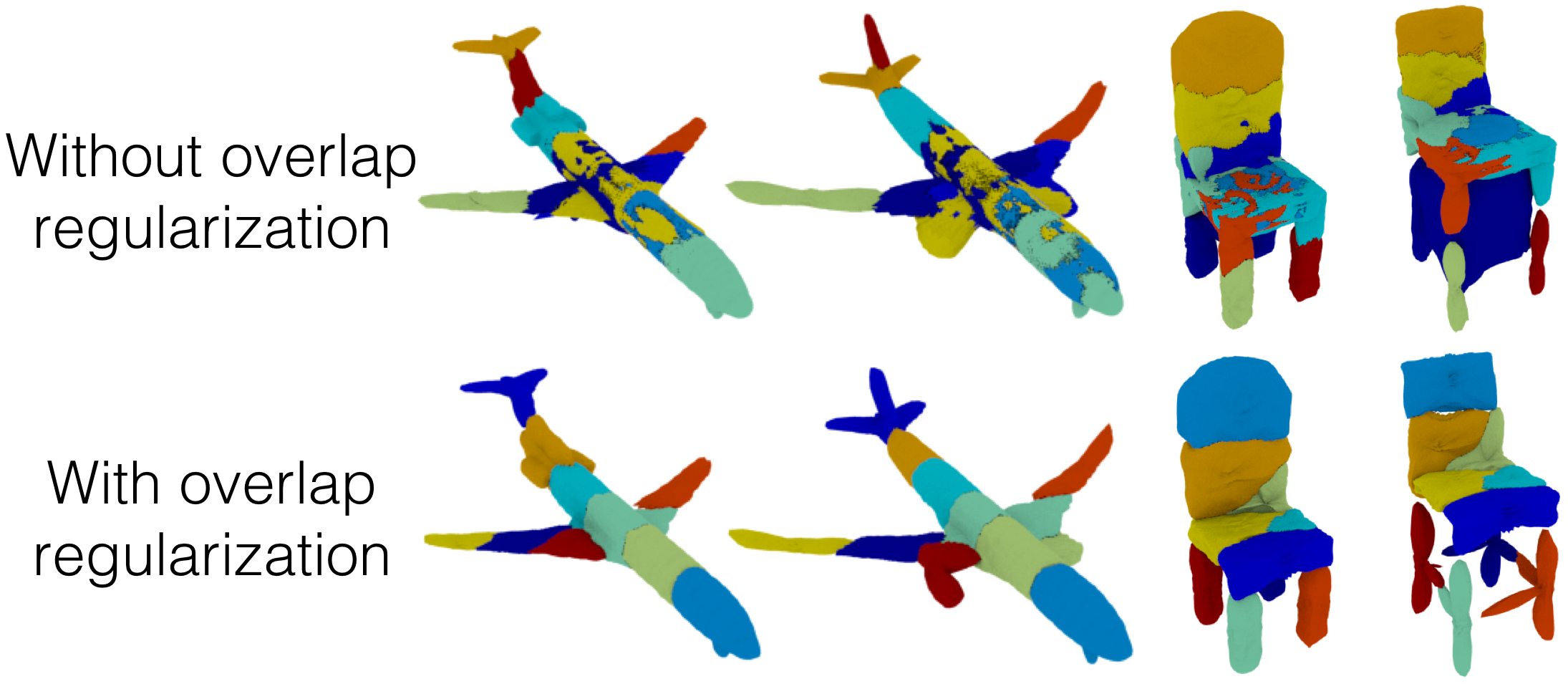}

\vspace{-0.5\baselineskip}
\caption{Visualizations of the effect of regularizing the overlap for the primitives decomposition result of the airplane and chair categories.}
\label{fig:overlap_vis}

\end{center}
\end{minipage}

\hspace{0.01\textwidth}

\begin{minipage}{0.47\textwidth}
\begin{center}

\makeatletter
\def\@captype{table}
\makeatother

\begin{tabular}{cccc}
\bhline{1.5 pt}

         & \multicolumn{3}{c}{airplane}        \\
         & Overlap    & F-score     & Label IoU    \\\hline 
w/o reg.  & 5.810      & 69.55       & 48.15       \\
w/ reg. & {\bf 0.445}& {\bf 69.92} & {\bf 50.90} \\\hline 
         & \multicolumn{3}{c}{chair}           \\
         & Overlap    & F-score     & Label IoU    \\\hline 
w/o reg.  & 51.21      & {\bf 35.56} & 56.12       \\
w/ reg. & {\bf 1.16} & 33.92       & {\bf 64.37} \\


\bhline{1.5 pt}
\end{tabular}

\caption{Effects of the overlap regularization. The overlap score is scaled by the value of 1000 from the original value.}

\label{tb:overlaptable}

\end{center}

\end{minipage}

\end{tabular}

\vspace{-1\baselineskip}

\end{figure}

We visualize the effect of the overlap regularization in Figure \ref{fig:overlap_vis}. In the decomposition results, there is less overlap between primitives with the overlap regularization. Quantitative evaluation is shown in Table \ref{tb:overlaptable}. In the table, we define an overlap metric (denoted as "overlap" in the table), which counts the number of 3D points inside more than one primitive. The definition can be found in the appendix. Applying the overlap regularizer clearly reduces the overlap with a slight change in the F-score, and it improves the part IoU for both categories. In particular, the part IoU for the chair category increased significantly by 8\%.

Note that planar mesh patch as primitives \cite{groueix2018papier, deprelle2019learning, bednarik2020} also have high expressivity and suffer from the same overlapping problems as NSD. Existing overlap regularization for this type of primitives, however, needs computationally expensive Jacobian computation  \cite{bednarik2020}. Moreover, it is an indirect overlap regularization. We demonstrate that by simultaneously being highly expressive and being an implicit representation, NSD allows for a computationally simpler and more direct approach to overcome this problem. 

\paragraph{Semantic part.}
In Figure \ref{fig:overview}, it can be seen that a single NSD primitive (in cyan color) reconstructs empennage. Moreover, in Figure \ref{fig:primitive_visualize}, wings (colored in green) and fuselage (colored in blue) are reconstructed with nacelles by a single primitive each. This shows that NSD is able to reconstruct complex shapes in a way that multiple parts under same semantic part are reconstructed by one primitive. This shows the expressive power of NSD in reconstructing semantic parts. 

\subsection{Mesh sampling}
\label{sec:mesh_gen}
As the proposed method is able to represent surfaces in explicit forms with mesh templates, it is much faster in sampling meshes compared to time consuming isosurface extraction methods. To show this, we evaluate the meshing speed and reconstruction accuracy of our explicit representation against an implicit representation using a leading isosurface extraction method MISE \cite{mescheder2019occupancy}. For comparison, we use the same NSDN model for both representations. The results are shown in Table \ref{tb:meshtable}. NSD is able to sample meshes with comparable F-scores much faster than MISE (see NSDN ico$\#$2 and MISE up$\#1$). We also investigate the number of vertices and faces on the surface over mesh sampling speeds. Our method is able to produce higher resolution meshes significantly faster than MISE. We also show mesh sampling speed of BSP-Net \cite{chen2019bsp} as a reference for the implicit representation approach with fast mesh sampling. Our method is comparable with \cite{chen2019bsp}. Note that we show the result of BSP-Net only for referring to the meshing speed and quality, as it focuses on low polymesh. 

\section{Conclusion}
In this study, we propose NSD as a novel primitive representation. We show that our method consistently outperforms previous primitive-based approaches and show that the only primitive-based approach performing relatively better than the leading reconstruction technique (OccNet \cite{mescheder2019occupancy}) in single view reconstruction task. We also show that our primitive based approach achieves significantly better semantic capability of reconstructed primitives. As a future work, we would like to integrate texture reconstruction to extend our primitive based approach for more semantic part reconstruction.

\subsubsection*{Broader impact}
A potential risk involved with NSD is that it can be extended to plagiarize 3D objects, such as furniture and appliance design etc. As our NSDN solely consists of very simple fully connected layers and mesh extraction processes, it is very fast and cost effective; one might be able to run our model on mobile devices with proper hardware optimization. This opens up more democratized 3D reconstruction, but it also comes with the possible risk of being applied to plagiarize the design of real world products by combination with 3D printers.

\subsubsection*{Acknowledgments}
We would like to thank Antonio Tejero de Pablos, Hirofumi Seo, Hiroharu Kato, James Daniel Borg, Kenzo-Ignacio Lobos-Tsunekawa, Ryohei Shimizu, Shunya Wakasugi, Yusuke Kurose and Yusuke Mori for insightful feedback. We also appreciate the members of the Machine Intelligence Laboratory for constructive discussion during the research meetings. This work was partially supported by JST AIP Acceleration Research Grant Number JPMJCR20U3, and partially supported by JSPS KAKENHI Grant Number JP19H01115.

\bibliography{egbib}

\appendix
\section{Example of cartesian spherical harmonics}
Spherical harmonics expansion $f_{\infty}$ with Cartesian spherical harmonics $Y_{l, m}$ is written as follows:
\begin{equation}
f_{\infty}({\bf d}) = \sum_{l=0}^{\infty}\sum_{m=-l}^{l} c_{l,m} Y_{l,m}(\omega({\bf d})),\ \omega({\bf d})=(\sin \theta \cos \phi, \sin \theta \sin \phi, \cos \theta)
\end{equation}
where ${\bf d} = (\theta, \phi)\in \mathbb{S}^2$, $c_{l, m}\in \mathbb{R}$ is a constant. A Few examples of $Y_{l, m}$ given $(l, m)$ are show below:
\begin{align*} 
Y_{0, 0}(x, y, z) &= \frac{1}{2}\sqrt{\frac{1}{\pi}} & & & & \\ 
Y_{1, -1}(x, y, z) &= \sqrt{\frac{3}{4\pi}}y \\
Y_{1, 0}(x, y, z) &= \sqrt{\frac{3}{4\pi}}z \\
Y_{1, 1}(x, y, z) &= \sqrt{\frac{3}{4\pi}}x \\  
Y_{2, -2}(x, y, z) &= \frac{1}{2}\sqrt{\frac{15}{\pi}}xy &
Y_{2, -1}(x, y, z) &= \frac{1}{2}\sqrt{\frac{15}{\pi}}yz \\
Y_{2, 0}(x, y, z) &= \frac{1}{4}\sqrt{\frac{5}{\pi}}(-x^2-y^2+2z^2) \\
Y_{2, 1}(x, y, z) &= \frac{1}{2}\sqrt{\frac{15}{\pi}}zx &
Y_{2, 2}(x, y, z) &= \frac{1}{4}\sqrt{\frac{15}{\pi}}(x^2-y^2)
\end{align*}

\section{Definition of $G$ and $G^{-1}$}
We define conversion from Cartesian coordinate to sphere surface $G:\mathbb{R}^3\rightarrow\mathbb{S}^2$ as:
\begin{equation}
G(x, y, z) = (\arctan\frac{y}{x}, \arctan\frac{\sqrt{x^2 + y^2}}{z^2})
\end{equation}
We define conversion from spherical coordinate to Cartesian coordinate $G^{-1}: {\mathbb R} \times {\mathbb S}^2 \rightarrow {\mathbb R}^3$ as:
\begin{equation}
G^{-1}(r, \theta, \phi) = (r\sin \theta \cos \phi, r\sin \theta \sin \phi, r\cos \theta)
\end{equation}

\section{Visualization of single view reconstruction}
\begin{figure}

  \centering
      \includegraphics[width=13cm]{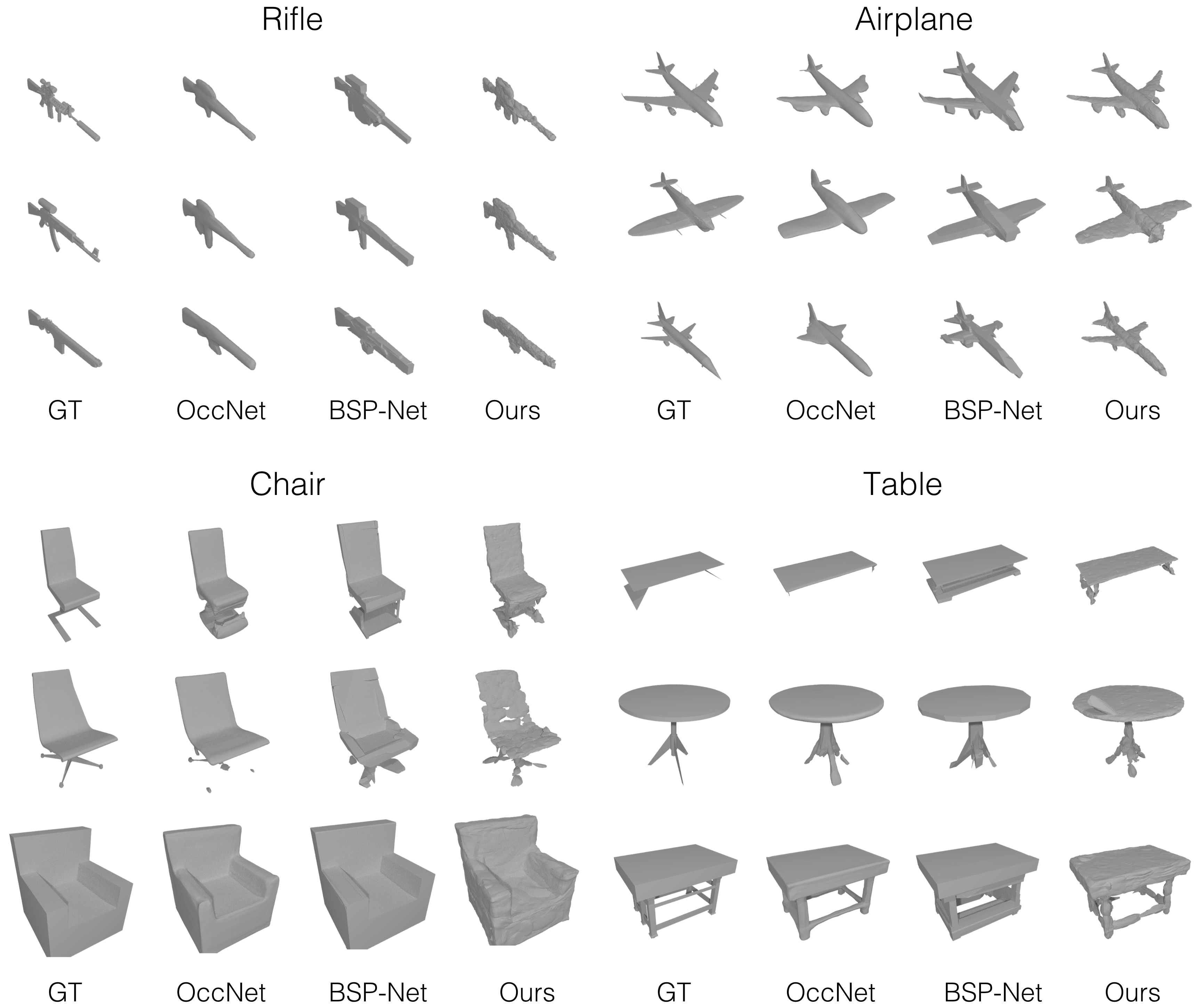}
              \caption{The additional visualization of the single view reconstruction results.}
        \vspace{-1\baselineskip}

\label{fig:svr_more}
\end{figure}

We show the additional visualization of the single view reconstruction results of rifle, airplane, chair and table categories from ShapeNet \cite{chang2015shapenet} in Figure \ref{fig:svr_more}.

\section{Visualization of primitives}
\begin{figure}

  \centering
      \includegraphics[width=12cm]{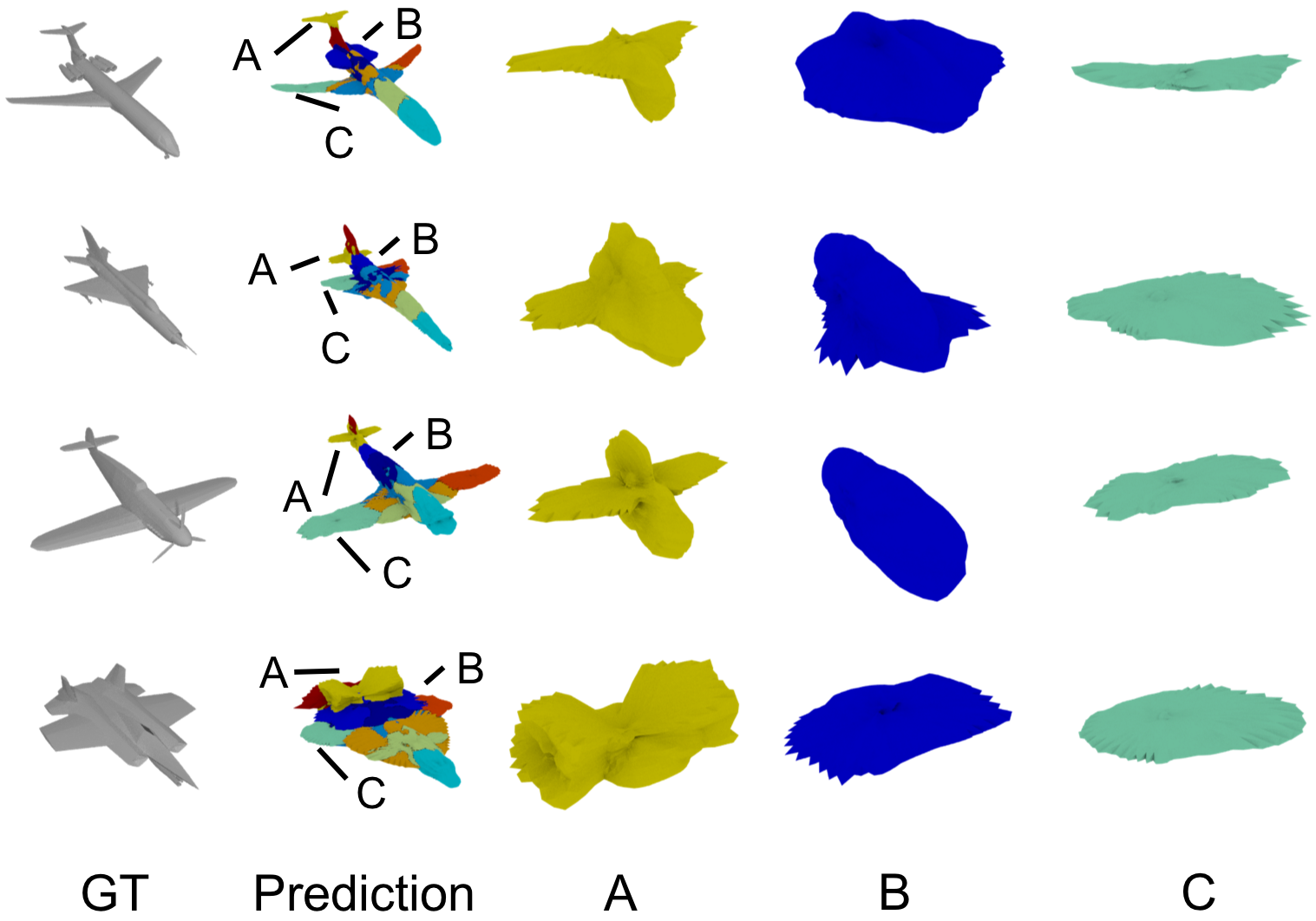}
              \caption{The additional visualization of our primitives of the airplane category.}
        \vspace{-1\baselineskip}

\label{fig:prim_plane}
\end{figure}

\begin{figure}

  \centering
      \includegraphics[width=12cm]{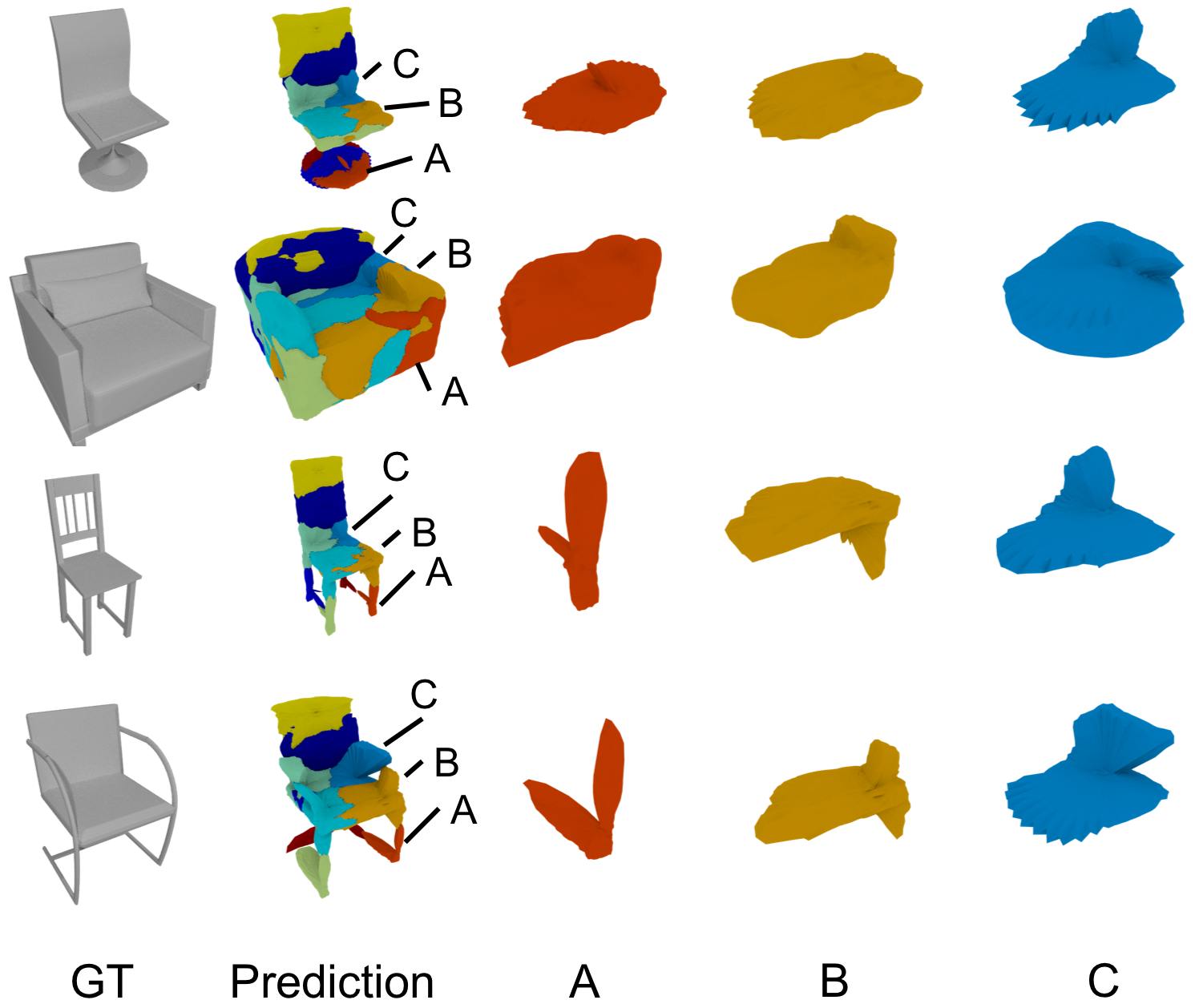}
              \caption{The additional visualization of our primitives of the chair category.}
        \vspace{-1\baselineskip}

\label{fig:prim_chair}
\end{figure}

\begin{figure}

  \centering
      \includegraphics[width=12cm]{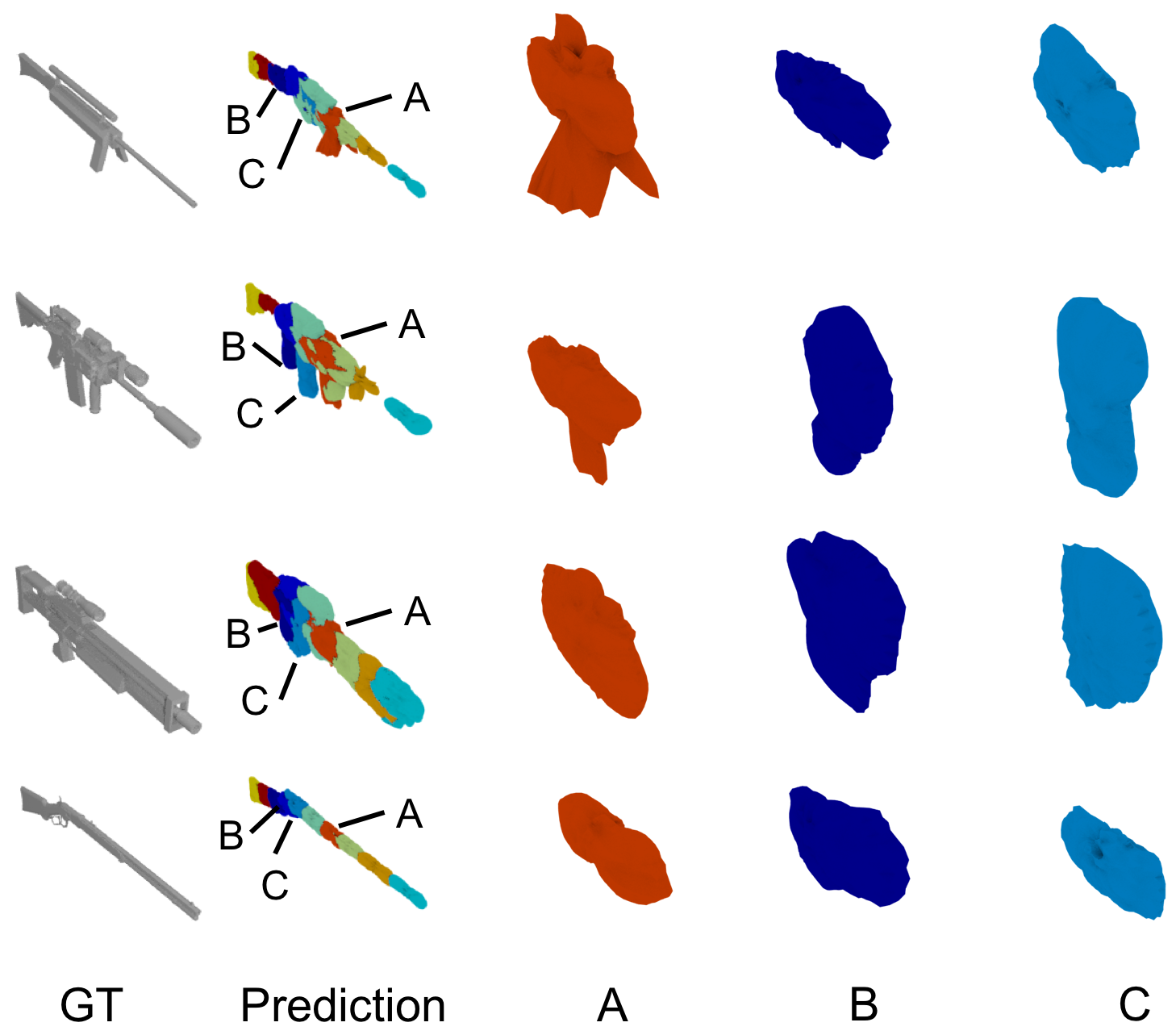}
              \caption{The additional visualization of our primitives of the rifle category.}
        \vspace{-1\baselineskip}

\label{fig:prim_rifle}
\end{figure}

We show the additional visualization of our primitives in Figure \ref{fig:prim_plane}, \ref{fig:prim_chair} and \ref{fig:prim_rifle} for plane, chair and rifle categories from ShapeNet \cite{chang2015shapenet}, respectively.

\begin{figure}

  \centering
      \includegraphics[width=12cm]{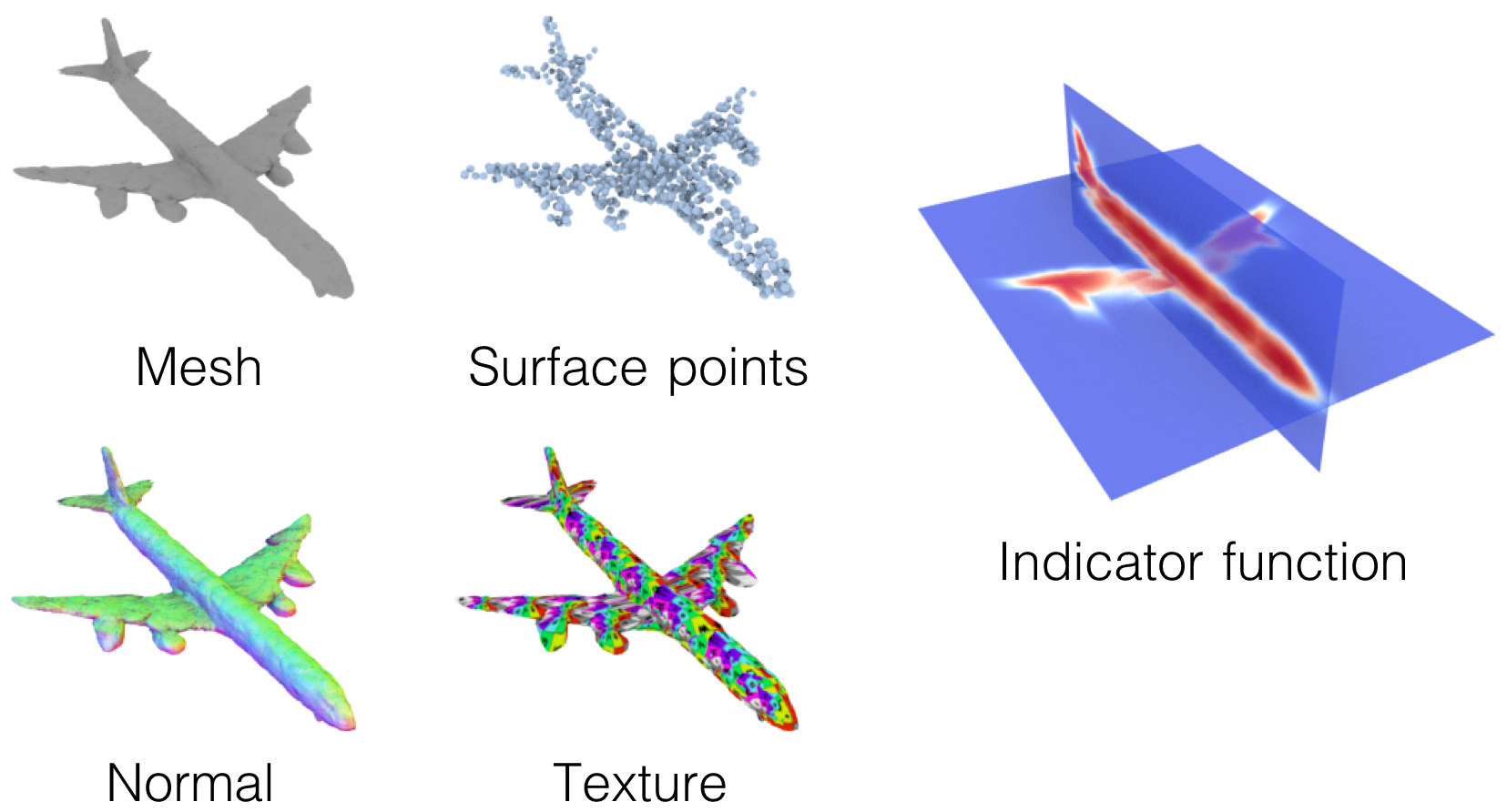}
              \caption{Differentiable shape and surface representations of NSD.}

\label{fig:repr_vis}
\end{figure}
\section{Visualization of differentiable shape and surface representations}
NSD provides multiple \emph{differentiable} shape and surface representations which are available both during training and inference: mesh, surface points, normal, indicator function (signed distance function) and texture. We visualize them in Figure \ref{fig:repr_vis}.
\paragraph{Normal estimation}
As shown in Figure \ref{fig:repr_vis}, NSD can also estimate differentiable normal vectors.
Unlike the methods using mesh templates, our approach is able to derive normal in arbitral resolution. Following \cite{mescheder2019occupancy}, a surface normal of the $i$ th primitive ${\hat n}_i$  can be derived as follows:
\begin{equation}
{\hat n}_i({\bf {\hat p}}; {\bf t}_i) = -\frac{\partial{\hat O}_i({\bf {\hat p}}; {\bf t}_i)}{\partial{\hat p}}
\end{equation}
where ${\bf {\hat p}} \in {\hat P}_i$ is a predicted surface point, ${\hat O}_i$ is an indicator function and ${\bf t}_i$ is a translation vector of the $i$ th primitive.
Collective surface normal vectors ${\hat n}$ can be defined as follows:
\begin{equation}
  {\hat n} =  \bigcup_{i}{\{{\hat n}_i({\bf {\hat p}}; {\bf t}_i)|\, \forall j \in [N\setminus i],\,  {\hat O_j}({\hat P}_i({\bf d}; {\bf t}_i); {\bf t}_i)<\tau_s,\, {\bf d} \in \{{\bf d}_k\}^K_{k=1}\}}
\end{equation}
where $N$ is a number of primitives and $\tau_s$ is a hyperparameter for the threshold of the isosurface indicator value.
Note that differentiable normal estimation during training by the above approach is possible by having implicit and explicit representations, whereas the approach of \cite{mescheder2019occupancy} can extract normal vectors only at inference time.

\section{Analysis on expressive power of primitive shapes}
\begin{figure}
        \vspace{+1\baselineskip}

  \centering
      \includegraphics[width=13cm]{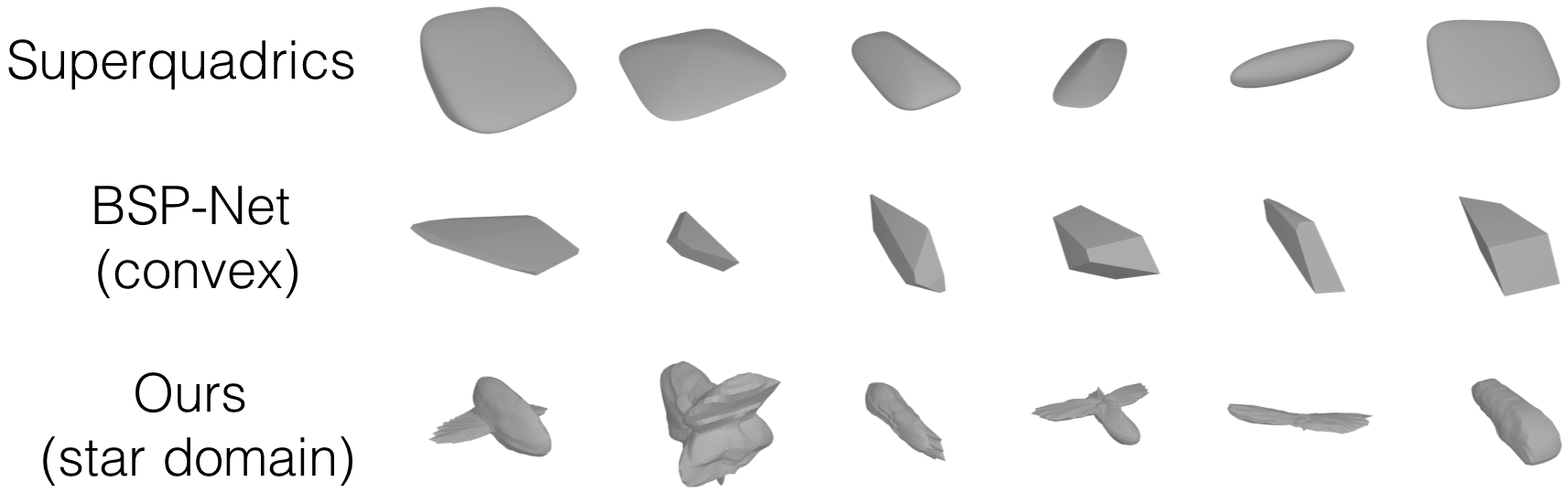}
              \caption{Randomly sampled primitives: superquadrics \cite{Paschalidou2019CVPR}, convex \cite{chen2019bsp} and ours (star domain).}

\label{fig:prim_each_vis}
\end{figure}

We quantitatively evaluate the expressive power of NSD against other primitives of the previous works: convexes \cite{chen2019bsp} and superquadrics \cite{Paschalidou2019CVPR}. We evaluate the expressive power by measuring the complexity of the inferred primitive shapes. To quantify the complexity of the shape, we evaluate discrete gaussian curvature \cite{cohen2003restricted}. We use the airplane and the chair categories from ShapeNet \cite{chang2015shapenet} in this evaluation. For NSD, we use $N = 10$ for the number of primitives. We show mean and standard deviation of the curvature measure in Table \ref{tb:curv}. The larger mean value indicates primitive shapes have more complex surfaces in terms of unevenness and the larger standard deviation means primitives have more diverse shapes. We see that NSD has the largest mean and standard deviation compared to the previous works. This quantitatively shows NSD has more expressive power as it learns more complex and diverse primitive shapes. We also visualize randomly sampled primitives from the airplane and chair categories in Figure \ref{fig:prim_each_vis}.

\begin{table}
\centering

\begin{tabular}{c|cc}
\bhline{1.5 pt}
& mean
& std
\\ \hline 
Superquadrics \cite{Paschalidou2019CVPR} &0.042 & 0.030 \\
BSP-Net (convex) \cite{chen2019bsp} &0.070  &0.344 \\ 
Ours (star domain) & {\bf 0.154} & {\bf 0.351} \\ 
\bhline{1.5 pt}
\end{tabular}
        \vspace{+1\baselineskip}
\caption{Mean and standard deviation of discrete gaussian curvature \cite{cohen2003restricted}.}
\label{tb:curv}
\end{table}
\section{Definition of the overlap regularizer}
We adapt the decomposition loss proposed in \cite{deng2019cvxnets} as an off-the-shelf overlap regularizer. Note that we use L1 norm instead of L2 norm in the equasion, which results in:
\begin{equation}
L_{\rm decomp}(\Theta) = \mathbb{E}_{{\bf x}\sim\mathbb{R}^3}|{\rm ReLU}(\sum_i{\hat O_i({\bf x}; {\bf t}_i)} - \tau_r)|
\end{equation}
$\tau_r$ is the hyperparameter which controls the amount of the overlap.
\section{Formulation of the overlap count}
We quantify primitive overlap by counting the number of 3D points inside more than one primitive as follow:
\begin{equation}
{\rm Overlap} = \mathbb{E}_{{\bf x}\sim\mathbb{R}^3}\mathbbm{1}(\sum_i\mathbbm{1}({\hat O}_i({\bf x} ; {\bf t}_i) \geq \tau_s) > 1) 
\end{equation}
$\mathbbm{1}$ is an indicator function.

\end{document}